\documentclass{IEEEtran}
\usepackage[hyphens]{url}
\usepackage{booktabs}
\usepackage{cite}
\usepackage[utf8]{inputenc}
\usepackage{enumitem}
\usepackage{multirow}
\usepackage[algoruled,linesnumbered]{algorithm2e}
\usepackage{pifont}
\usepackage{graphicx}
\usepackage{amsmath}
\usepackage{authblk}
\usepackage[hidelinks]{hyperref}
\usepackage[noabbrev]{cleveref}

\newcommand{\cmark}{\ding{51}}%
\newcommand{\xmark}{\ding{55}}

\newlist{inlinelist}{enumerate*}{1}
\setlist*[inlinelist,1]{%
  label=(\arabic*),
}

\begin{document}

\title{CharBot: A Simple and Effective Method for Evading DGA Classifiers}
\author[1,2]{Jonathan Peck}
\author[3]{Claire Nie}
\author[3]{Raaghavi Sivaguru}
\author[3]{Charles Grumer}
\author[4]{Femi Olumofin}
\author[4]{Bin Yu}
\author[3]{Anderson Nascimento}
\author[1,3]{Martine De Cock}

\affil[1]{Department of Applied Mathematics, Computer Science and Statistics, Ghent University, Ghent, 9000, Belgium}
\affil[2]{Data Mining and Modeling for Biomedicine, VIB Inflammation Research Center, Ghent, 9052, Belgium}
\affil[3]{School of Engineering and Technology, University of Washington, Tacoma, WA 98402, USA}
\affil[4]{Infoblox, Santa Clara, CA 95054, USA}

\maketitle

\begin{abstract}
Domain generation algorithms (DGAs) are commonly leveraged by malware to create lists of domain names which can be used for command and control (C\&C) purposes. Approaches based on machine learning have recently been developed to automatically detect generated domain names in real-time. In this work, we present a novel DGA called \textit{CharBot} which is capable of producing large numbers of unregistered domain names that are not detected by state-of-the-art classifiers for real-time detection of DGAs, including the recently published methods FANCI (a random forest based on human-engineered features) and LSTM.MI (a deep learning approach). CharBot is very simple, effective and requires no knowledge of the targeted DGA classifiers. We show that retraining the classifiers on CharBot samples is not a viable defense strategy. We believe these findings show that DGA classifiers are inherently vulnerable to adversarial attacks if they rely only on the domain name string to make a decision. Designing a robust DGA classifier may, therefore, necessitate the use of additional information besides the domain name alone. To the best of our knowledge, CharBot is the simplest and most efficient black-box adversarial attack against DGA classifiers proposed to date.
\end{abstract}

\section{Introduction}
The purpose of distributing malware is often to extract sensitive information from victim machines or to use them for disseminating spam. To achieve this, botmasters need to be able to communicate with the infected machines, which is done via command-and-control (C\&C) servers. The use of a fixed pool of C\&C servers is not attractive, however, since these servers may be taken offline or blacklisted. Therefore, malware authors design \textit{domain generation algorithms} or DGAs to automatically create many domain names that are likely to be unregistered and hence available for the malware to establish a communication channel~\cite{plohmann2016comprehensive}. A DGA makes use of a \textit{seed}, i.e., some random number that is accessible to both the botmaster and the malware on the infected machines.

Possible seeds include the current date, trending topics on Twitter, weather forecasts, etc. Once this seed has been fixed, the botmaster, as well as all of the infected machines, can generate the same list of domains. The botmaster registers one of these domains and waits for the malware to successfully resolve a DNS query against it. From that point on, communication can take place. Should the C\&C server ever be taken offline or have its domain blacklisted, this process can simply be restarted and a new C\&C server can be established.

An extensive amount of research in the past decade has been devoted to the development of methods for detection of domains generated by DGAs~\cite{antonakakis2012throw,yadav2012detecting,Woodbridge2016,tran2018lstm}. These methods can be roughly divided into two classes:
\begin{inlinelist}
    \item classifiers that detect DGAs based solely on the domain name itself; and
    \item classifiers that use some sort of context information,  such as IP addresses of the source, traffic, and query patterns by the infected machines.
\end{inlinelist}
Our focus in this paper is on the first kind of classifiers, i.e.~techniques that can detect DGA domains in real-time based on the domain name string. These systems are particularly attractive since additional information beyond the domain name string can be expensive to acquire. It might also simply not be available due to privacy concerns.
Another significant advantage of systems that perform DGA detection based solely on the domain name is their potential use in real-time systems, blocking malicious domains before they are actually resolved.  
Accordingly, much research has been carried out to prevent this type of C\&C communication using systems that can detect in real-time whether a domain name is likely generated by a DGA or not~\cite{antonakakis2012throw,Woodbridge2016,lison2017automatic,Yu:2017a,cchoudhary2018a,rhodes2018,pereira2018a,schuppen2018,rsivaguru2018,tran2018lstm,Yu2018a}. 

Such DGA classifiers need to be sufficiently robust so that they can still reliably detect DGA domains even when the DGAs start generating lists from seeds that were not seen during training. Existing work in this area is comprised both of methods that make use of human-engineered features as well as deep learning techniques which learn to extract relevant features automatically. We show in this paper that both kinds of methods are inherently vulnerable to simple attacks and hence the use of side information may be crucial to developing robust DGA classifiers. Specifically, we introduce a new and effective DGA called \textit{CharBot}. It is a simplistic character-based DGA (hence the name) that generates domain names by randomly modifying two characters in well known benign domains collected from the Alexa top domain names.\footnote{\url{https://alexa.com/topsites}. Accessed: 2019-02-10.} We find that the domains CharBot generates are almost always unregistered, hence available for C\&C communication.

To demonstrate CharBot's capabilities, we attack two types of recently proposed prototypical DGA classifiers that are considered state-of-the-art at the time of this writing:
\begin{inlinelist}
    \item a random forest (RF) model called \textit{FANCI} based on human-engineered features extracted from the domain name~\cite{schuppen2018} and
    \item a deep neural network (DNN) model called \textit{LSTM.MI}~\cite{tran2018lstm}.
\end{inlinelist}
We also test a RF approach called \textit{B-RF} based on the features proposed in~\cite{rsivaguru2018}. We train these models on data sets consisting of benign and malicious domain names. The benign names originate from the Alexa top domain names. For the malicious domains, we use the OSINT Bambenek Consulting feeds.\footnote{\url{http://osint.bambenekconsulting.com/feeds/}. Accessed: 2019-02-10.} We find that the domain names generated by CharBot go largely undetected by all these state-of-the-art DGA classifiers. 

We attempt to harden the classifiers against CharBot by incorporating samples from it in the training data sets and retraining the models. Although this strategy does increase the detection rates, they are still not high enough to be practical. We also try retraining using samples generated by DeepDGA --- a state of the art generative model for malicious domain names~\cite{anderson2016deepdga} --- as well as the DeceptionDGA by Spooren et al.~\cite{spooren2019detection}, but we find that this does not adequately help with detecting CharBot. CharBot is much simpler than both DeepDGA and DeceptionDGA: DeepDGA is a deep learning approach, whereas CharBot performs only simple string manipulations; DeceptionDGA is designed to evade classifiers based on human-engineered features. By contrast, CharBot is fully black-box: it does not require any details of the models being attacked.

CharBot works by corrupting domain names from the Alexa top domains, so it is natural to ask whether the domains it generates can also be used to successfully attack DGA classifiers that do not depend on Alexa for training. To answer this question, we investigate whether the DGA classifiers can be hardened by replacing the Alexa data set by an alternative data set of benign domains during training. To this end, we use a data set of domain names that occurred in real DNS traffic, weakly labeled according to heuristic rules \cite{Yu:2017a}. We find that training on this different data set yields approximately the same results as when training on Alexa. This supports the idea that CharBot attacks are transferable across models and data sets.

These findings expose a dangerous weakness in modern DGA classifiers: they can be circumvented using a simple algorithm and they cannot be easily trained to detect it well. We speculate that this weakness is inherent in any model that relies solely on domain name strings to perform DGA classification. CharBot works by introducing a small number of typographical errors in benign domain names from the Alexa data set. As such, the statistical properties of the names it generates will be almost identical to those of the Alexa domains. This makes it nearly impossible for a classifier to draw any significant distinction between Alexa names and CharBot names. Moreover, any other set of legitimate domains that should be accepted by a classifier with high probability could in principle be used instead of Alexa by a CharBot attack. Therefore, we do not believe these attacks can be mitigated without relying on additional side information. Such information might include the IP addresses the domains resolve to, how many times the domains were queried and when, etc. This has been explored in other works already~\cite{yadav2012detecting,schiavoni2014phoenix,kwon2016psybog,lison2017neural,pereira2018a}. To our knowledge, we are the first to expose this type of weakness in DGA classifiers that do not use side information. We would, therefore, recommend that the community focuses its research efforts on DGA classifiers that utilize side information and not just rely on the domain name string by itself.

The rest of this paper is structured as follows. \Cref{sec:related} gives an overview of related work in the field of adversarial machine learning. \Cref{sec:charbot} details the CharBot algorithm. \Cref{sec:data} describes the data sets we used for the experiments. \Cref{sec:experiments} outlines our experiments and discusses their results, as well as several ways we could defend against CharBot attacks. \Cref{sec:conclusion} concludes the work and lists some possibilities for future research.

\section{Related work}\label{sec:related}
Machine learning approaches that leverage the domain name string for DGA detection can be categorized into two groups:
\begin{inlinelist}
    \item so-called ``featureful'' methods that rely on human defined lexical features extracted from the domain names, such as domain name length, vowel-character ratio, bigrams, etc.~\cite{antonakakis2012throw,schiavoni2014phoenix,schuppen2018} and
    \item ``featureless'' methods in which the automatic discovery of good features is part of the overall machine learning model training process, as a form of representation learning~\cite{Woodbridge2016, lison2017automatic,Yu:2017a,rhodes2018,tran2018lstm,Saxe2017}.
\end{inlinelist}
Popular kinds of classifiers used in the featureful approach for DGA detection are logistic regression and tree ensemble methods, while the featureless approach relies on the use of deep neural networks, namely Long Short-Term Memory (LSTM) networks and Convolutional Neural Networks (CNN). Most papers about the featureless approach include a featureful approach as a baseline method \cite{Woodbridge2016,Yu2018a,Yu:2017a,lison2017automatic,cchoudhary2018a,rsivaguru2018,tran2018lstm,Saxe2017}, and the featureless approach is typically reported to yield better, more accurate results.

A natural response of malware authors to machine learning classifiers for DGA detection is to try to purposely craft domain names that will be mislabeled as benign by the classifiers. This kind of evasion attack is studied as part of the broader field of \textit{adversarial machine learning} (AML)~\cite{vorobeychik2018adversarial}. In this setting, an intelligent adversary aims to exploit weaknesses in a machine learning model in order to obtain desired (illegitimate) outcomes. A prototypical example is that of spam classification, where the adversary attempts to craft spam e-mails that evade detectors while still achieving the desired results. Seminal contributions in this area include the work of Dalvi et al.~\cite{dalvi2004adversarial} as well as the papers by Lowd and Meek~\cite{lowd2005adversarial,lowd2005good}, who study classical machine learning algorithms such as linear classifiers, naive Bayes, support vector machines and maximum entropy filters. More recent works primarily study AML for deep neural networks~\cite{goodfellow2014explaining,madry2017towards,raghunathan2018semidefinite}.

A recent innovation in the area of deep learning and generative modeling,  is the \textit{Generative Adversarial Network} or GAN, first proposed by Goodfellow et al.~\cite{goodfellow2014generative}. In the GAN framework, a generative model is trained by pitting it against an adversary. The adversary is a discriminative model whose goal is to discern whether a given sample came from the data generating distribution or from the generative model. The generator is trained to maximize the loss of the discriminator, so the GAN training procedure corresponds to a two-player minimax game. Ideally, when the training converges, the generator should recover the data generating distribution and the discriminator should not be able to do any better than random guessing.

GANs have found several uses in cybersecurity by now. Anderson et al.~\cite{anderson2016deepdga} proposed \textit{DeepDGA}, which is a generative model for DGA domains trained using a GAN. They find that adding samples from DeepDGA to the training data of deep learning based DGA classifiers improves their performance against unseen malware families, aiding generalization of the models when insufficient training data is available. In the field of password security, Hitaj et al.~\cite{hitaj2017passgan} have proposed \textit{PassGAN}, another generative model trained in the GAN framework. PassGAN learns to capture the distribution of human passwords and is able to surpass state of the art tools for password guessing. Hu and Tan~\cite{hu2017generating} recently proposed \textit{MalGAN}, a GAN with which they are able to construct malware samples that can bypass black-box machine learning methods. Their attack is particularly striking because they are able to reduce malware detection rates to almost zero without requiring direct access to the detectors they aim to evade. Moreover, they found that explicitly retraining the detectors on MalGAN samples is ineffective: MalGAN can easily be adapted to take this retraining into account, bypassing the retrained models again with almost 100\% success. With CharBot, we achieve similar (and, in several cases, better) results with a much simpler approach that can actually be incorporated within a piece of malware, in contrast to deep-learning based methods which are usually too large or too computationally intensive.

Several authors have recently looked into the automatic generation of URLs for phishing.
To this end, Bahnsen et al.~\cite{bahnsendeepphish} create a text consisting of known phishing URLs from PhishTank\footnote{\url{https://www.phishtank.com/}. Accessed: 2019-02-08.} and use it to train an LSTM for text generation, i.e.~given a small seed sentence, predict the next characters iteratively. They report that this technique generates examples that are not detected by their own LSTM based phishing URL classifier~\cite{bahnsen2017classifying}. 
Anand et al.~\cite{anand2018} trained a GAN -- containing a character based LSTM as part of its architecture -- to generate synthetic phishing URLs to augment the training data for feature-based phishing URL detection classifiers. The problem they address is the class imbalance in typical training data sets, which contain many more examples of benign URLs than of phishing URLs. Instead of adding all generated phishing URLs as positive examples to their training data, they first map the generated URLs to their corresponding feature vectors, and select ``representative samples'' based on Euclidean distance in this feature space.
In a similar vein to Anand et al., Burns et al.~\cite{burns2019} train a GAN on OpenPhish\footnote{\url{https://openphish.com/}. Accessed: 2019-02-08.}, PhishTank and DNS-BH\footnote{\url{http://www.malwaredomains.com/}. Accessed: 2019-02-08.} data sources to develop synthetic phishing domains. They compare a random forest classifier trained on Alexa and Umbrella\footnote{\url{http://s3-us-west-1.amazonaws.com/umbrella-static/index.html}. Accessed: 2019-02-08.} data sets to models that were augmented with samples generated by the GAN. They find that the augmented models appear to have consistently higher test set accuracy than the original classifier.

URLs intended for phishing are quite different in nature than DGA domains for C\&C purposes. Indeed, to be successful, phishing URLs need to deceive humans, which requires them to be as indistinguishable as possible from benign URLs to the human observer. DGA domain names used for C\&C purposes are not intended at all to be read by human users. DGA domain names are successful if they can evade DGA classifiers and have not been previously registered, i.e.~they should be available for the botmaster to register. To the best of our knowledge, so far Anderson et al.~are the only ones who have looked into generative modeling of DGA domain names~\cite{anderson2016deepdga}. Although their results are significant, we show in this work that classifiers which have been adversarially trained using DeepDGA remain vulnerable to simple attacks such as the CharBot algorithm we propose in \cref{sec:charbot}.

The CharBot algorithm is a \textit{black-box targeted evasion attack} that works against tree ensembles and neural networks. ``Black-box'' refers to the fact that our CharBot DGA does not require details of the classifiers in order to work: it can attack any model trained on any data set and succeed with high probability. It is a targeted attack because we want the classifiers to output a specific class in response to our DGA samples, namely \textit{benign}. Untargeted attacks, on the other hand, merely aim to change the classification to \textit{any} class other than the original; for example, an untargeted attack would also count a change from benign to malicious as a success, whereas in our scenario that would be unacceptable. Finally, CharBot is an evasion attack because it occurs at test time, when the model is already trained and deployed. This is in contrast to \textit{poisoning attacks} which occur at training time and work by corrupting samples in the training data set in order to deliberately introduce weaknesses into the model~\cite{vorobeychik2018adversarial}. The CharBot attack itself is inspired by \textit{typosquatting}, a well-known technique used by phishers and social engineers~\cite{szurdi2014long}. Typosquatting involves taking a legitimate domain and introducing a few typographical errors that are unlikely to be noticed by human users (e.g., changing \textit{google.com} into \textit{g0ogle.com}). Whereas \textit{a priori} one would think that typosquatting exploits an inherently human vulnerability, we make the surprising discovery that state-of-the-art DGA classifiers are actually vulnerable to such techniques as well.

Similarly to our work here, Spooren et al.~\cite{spooren2019detection} developed \textit{DeceptionDGA}, a novel DGA which incorporates knowledge of the features used by a DGA classifier in order to attack it. They report significant reductions in predictive accuracy for the FANCI model as well as the Endgame LSTM by Woodbridge et al.~\cite{Woodbridge2016}. The DeceptionDGA algorithm is more complicated than CharBot, requiring knowledge of the underlying model in order to deploy it. Despite this difference in complexity, the detection rates we observe for CharBot in our experiments are comparable to those of DeceptionDGA.

We also wish to acknowledge the concurrent work of~\cite{sidi2019maskdga} who describe \textit{MaskDGA}, a black-box technique for evading DGA classifiers that is similar to CharBot. MaskDGA makes use of a surrogate model as well as a list of DGA domains. It uses these data to craft character-level perturbations of the malicious domains such that they are no longer recognized by the surrogate model. Similarly to our own results, the authors find that such techniques are highly effective at reducing the accuracy of state-of-the-art DGA classifiers. They also make the recommendation that DGA classifiers should rely on additional side-information whenever this is possible in order to mitigate adversarial attacks. 

\section{CharBot}\label{sec:charbot}
CharBot is a character-based DGA intended to show how successful a simplistic DGA based on small perturbations can be at evading detection by state-of-the-art classifiers. Without loss of generality, throughout this paper we consider domains consisting of a second-level domain (SLD) and a top-level domain (TLD), separated by a dot, as in e.g.~\textit{wikipedia.org}. 
CharBot requires the following inputs:
\begin{itemize}
    \item A list of legitimate domain names. In our case, ten thousand Alexa domains with a second-level domain (SLD) length of six or greater are used.
    \item A list of top-level domains (TLDs).
    \item A date to be used as a seed for pseudorandomization.
\end{itemize}
With these inputs, CharBot (1) selects a domain from the provided list, (2) selects two characters from the SLD, and (3) selects two replacement characters. The replacement characters are chosen from an equal distribution of DNS-valid characters --- the alphanumeric characters and the dash --- and the algorithm ensures the characters selected from the SLD are different from the replacement characters. Finally, CharBot (4) appends a TLD to the new domain by selecting one of the following: \textit{com, at, uk, pl, be, biz, co, jp, cz, de, eu, fr, info, it, ru, lv, me, name, net, nz, org, us}. Pseudocode is given in \cref{alg:charbot}.

\begin{algorithm}
    \KwData{a list of SLDs $D$, a list of TLDs $T$, a seed $s$}
    \KwResult{a DGA domain}
    
    Initialize the pseudorandom generator with the seed $s$.\\
    Randomly select a SLD $d$ from $D$.\\
    Randomly select two indices $i$ and $j$ so that $1 \leq i,j \leq |d|$.\\
    Randomly select two replacement characters $c_1$ and $c_2$ from the set of DNS-valid characters.\\
    Set $d[i] \gets c_1$ and $d[j] \gets c_2$.\\
    Randomly select a TLD $t$ from $T$.\\
    \Return{$d.t$}
    \caption{CharBot}
    \label{alg:charbot}
\end{algorithm}

A DGA is successful if it can generate many unique domains that have not yet been registered and which are not flagged by DGA classifiers as malicious. CharBot draws its replacement characters from a uniform distribution. Therefore, the more characters we replace, the more the generated domains resemble random strings. This increases the detection rate by DGA classifiers, so we aim to keep the number of replacement characters minimal. We tested several choices for the number of characters to be replaced. We found that two characters strike an appropriate balance between the rate of detection by DGA classifiers and the probability that a domain is already registered: with two characters, domains are flagged slightly more often but almost all domains are unregistered (see \cref{tab:adversarialdata}); when replacing only a single character, detection rates go down but more domains turn out to be registered already.

Adversarial attacks such as CharBot are always accompanied by an \textit{adversarial cost function} $c(x, \tilde{x})$ which describes the cost associated with perturbing an ``ideal'' sample $x$ into a sample $\tilde{x}$ that the adversary can actually use. For image classification, it is common to use $\ell_p$ distances for this purpose~\cite{goodfellow2014explaining,raghunathan2018semidefinite,moosavi2016deepfool}. However, in our context, the cost of perturbing a correctly classified benign domain $x$ into a malicious domain $\tilde{x}$ that is classified as benign must be measured differently, as we are working in a discrete input space (text) instead of a continuous one (images). Specifically, it makes sense to define our cost function as follows:
\begin{align*}
    c(x, \tilde{x}) =
    \begin{cases}
        d_L(x, \tilde{x}) & \mbox{if $\tilde{x}$ is unregistered,}\\
        \infty & \mbox{otherwise.}
    \end{cases}
\end{align*}
Here, $d_L$ denotes the \textit{Levenshtein distance} or \textit{edit distance}~\cite{levenshtein1966binary}. The cost function $c(x,\tilde{x})$ increases with the number of edits (insertions, deletions, and substitutions) required to transform $x$ into $\tilde{x}$, as each edit makes the attack more detectable by DGA classifiers. However, there is an infinitely large cost associated with generating a domain that is already registered, since such domains cannot be used by the attacker at all and may cause the malware to malfunction. CharBot was designed to minimize this cost function efficiently and as simply as possible.

We note that obvious extensions of the CharBot algorithm are possible. For example, we could additionally implement insertions of new characters and deletions of existing characters. The number of characters might also be chosen adaptively based on some heuristic instead of fixed in advance. However, we chose to limit ourselves to only substituting a fixed number of characters since this simple strategy already gives us very good results. Moreover, simpler attacks are likely to be preferred by attackers and therefore constitute a greater security concern.

The only obstacle to the deployment of CharBot in real malware might be its size. We implemented CharBot in 17,983 bytes of Python code. The Alexa data set it requires takes up 145,008 additional bytes, although the public availability of this data set means it could be downloaded on the fly. Therefore, we would need at most 162,991 bytes for a full implementation of CharBot with Alexa included. By comparison, the DeepDGA algorithm \cite{anderson2016deepdga} requires to embed in the malware a trained machine learning model that takes up at least 6,539,192 bytes. This is about 40$\times$ larger than CharBot. We therefore feel that file size is no obstacle to deploying CharBot in real malware.

\section{Data sets}\label{sec:data}
We use three different kinds of data in our experiments:
\paragraph{Alexa} The top 1 million unique domain names from Alexa\footnote{\url{https://www.alexa.com}, Accessed 2019-02-08.}. Alexa ranks websites based on their popularity in terms of the number of page views and number of unique visitors. It only retains the websites' SLD and TLD, aggregating across any subdomains. For example, according to Alexa, the five highest ranked domain names in terms of popularity on 2019-02-06 are \textit{google.com}, \textit{youtube.com}, \textit{facebook.com}, \textit{baidu.com} and \textit{wikipedia.org}. It is generally assumed that the top 1 million domain names in the Alexa ranking are ``benign'' domain names in the sense that they were not created by a DGA. Of course, this does not mean that the domain is ``benign'' in the sense of not being used for malicious activity. Indeed, there is reason to believe that a significant number of Alexa top ranked domains are used for malicious purposes~\cite{royal2012quantifying}, but this is not the problem we are considering here. In our setting, we only consider a domain to be ``malicious'' if it was generated by a DGA.

\paragraph{Bambenek} 1 million unique DGA domain names from the Bambenek Consulting feeds\footnote{\url{http://osint.bambenekconsulting.com/feeds/}, Accessed 2019-02-08} for 3 different days, namely Jun 24, Jul 22, Jul 23, 2017. These feeds contain DGA domain names from specific malware families that were observed in real traffic on those days. Such domain names can be collected by reverse engineering a known malware family, generating lists of domain names with the reverse engineered malware, and checking which of these domain names also occur in real traffic.

\paragraph{Qname} 1 million unique domain names originating from a real-time stream of passive DNS data that consists of roughly 10-12 billion DNS queries per day collected from subscribers including ISPs (Internet Service Providers), schools, and businesses. We annotated this stream based on a set of heuristic filtering rules following~\cite{Yu:2017a}. Specifically, we labeled as benign all domains that
\begin{inlinelist}
    \item have been resolved at least twice,
    \item never resulted in an NXDomain response and
    \item span more than 30 days. Here, span is defined as the number of days between the first and last successfully resolved query for a given domain.
\end{inlinelist}
We randomly sampled 1 million such domains that appeared in DNS traffic
between September 2015 and August 2018. This data set is \textit{weakly labeled} since the heuristic filtering rules do not guarantee that the domains are actually benign or malicious; however, we believe it to be a useful approximation.\\ 

The Alexa and Qname data sets serve as our negative (benign) examples, whereas Bambenek serves as our set of positive (malicious) examples. 
Alexa and Qname have precisely 537 domains in common, which is a negligible number compared to the total sizes of the data sets, therefore making Qname a good data set to test transferability of CharBot.

We refer to the combination of Alexa and Bambenek data as \textit{AlexaBamb} and similarly for \textit{QnameBamb}. These data sets consist of 2 million samples each, 1 million per class. 


\section{Experiments}\label{sec:experiments}

We perform experiments on two DGA classifiers that are considered state of the art at the time of this writing: FANCI~\cite{schuppen2018} and LSTM.MI~\cite{tran2018lstm}, as well as a third model we call B-RF based on the work by~\cite{rsivaguru2018}. All classifiers are trained to label a domain name as either benign (negative) or malicious (positive). We find that the best results overall are achieved with the deep learning based LSTM.MI approach, followed by the random forest-based B-RF approach, and finally the random forest-based FANCI method. The difference in predictive accuracy between the various approaches is substantial. The results hold across the AlexaBamb and QnameBamb data sets (see table \ref{tab:results}, and a more detailed discussion in section \ref{sec:results}). 

To arrive at the results, we train on the AlexaBamb data set as well as on the QnameBamb data set with a 80\%/20\% train/test split for each, reporting the true positive rate (TPR), the partial area under the ROC curve (AUC) and the fraction of samples from CharBot, DeepDGA and DeceptionDGA which the models were able to detect (see table \ref{tab:results} and table \ref{tab:rates}). All of these metrics are reported at FPRs of 0.1\% and 1\%.\footnote{A low false positive rate is very important in deployed DGA detection systems because blocking legitimate traffic is highly undesirable. The threshold of 0.1\% FPR was chosen because this rate is often used by real-world models in practice, whereas 1\% is the largest FPR that could still be useful.} 
The AUC@0.1\%FPR is the integral of the ROC curve from FPR = 0 to FPR = 0.001 on the test data, and similarly for the AUC@1\%FPR. We repeat all experiments on the original models as well as the models after adversarial retraining.

To perform the adversarial retraining, we utilized the data sets shown in \cref{tab:adversarialdata}. Specifically, we used CharBot and DeepDGA with different seeds to generate training and testing data sets. The training data sets were used to augment the original training data of the classifiers; the testing data sets were used to verify their performance. For DeceptionDGA, Spooren et al.~\cite{spooren2019detection} supplied a list of 150,000 domains generated by their algorithm from which we sampled our training and testing data. Note that, based on a random sample of 500 domains\footnote{We limited ourselves to a random sample of 500 domains to avoid getting blocked by ISPs.}, CharBot has the highest fraction of unregistered domains (100\%), followed by DeepDGA (99.8\%) and DeceptionDGA (98.8\%).

The experiments on the QnameBamb data set are intended to investigate the transferability of CharBot. All CharBot domain names used in the experiments (see \cref{tab:adversarialdata}) are created by CharBot by corrupting domain names from the Alexa data set. This might leave DGA classifiers that are trained on AlexaBamb extra vulnerable to CharBot attacks.
A natural question to ask is whether CharBot can also successfully bypass DGA classifiers that were trained on a data set different from Alexa, one CharBot has no access to. To test this, we trained LSTM.MI, FANCI, and B-RF on the QnameBamb data and reported the same statistics as for AlexaBamb. 

Below we give a brief description of the LSTM.MI, FANCI, and B-RF classifiers, followed by detailed results (section \ref{sec:results}) and a discussion of possible countermeasures for defending against small perturbations attacks such as CharBot (section \ref{sec:counter}).

\begin{table*}
    \centering\begin{tabular}{lllrr}
        \toprule
        DGA & Data Set & Seeds Used & \# Unique Synthetic Domains & \# Unregistered Domains (out of 500 sampled)\\
        \midrule
        \multirow{2}{*}{CharBot}
        & Training & 2018-12-04 & 100,000 & 500 (100\%) \\
        & Testing & 2019-01-01 & 10,000 & 500 (100\%) \\
        \midrule
        \multirow{2}{*}{DeepDGA}
        & Training & 2018-12-04 & 100,000 & 499 (99.8\%) \\
        & Testing & 2019-01-01 & 10,000 & 499 (99.8\%) \\
        \midrule
        \multirow{2}{*}{DeceptionDGA}
        & Training & N/A & 100,000 & 494 (98.8\%)\\
        & Testing & N/A & 10,000 & 494 (98.8\%)\\
        \bottomrule
    \end{tabular}
    \caption{Adversarial data sets}
    \label{tab:adversarialdata}
\end{table*}

\begin{table}
    \centering\begin{tabular}{llcc}
        \toprule
        \# & Feature & FANCI & B-RF  \\
        \midrule
        1 & Domain name length & \cmark & \cmark \\
        2 & Second level domain length & \xmark & \cmark \\
        3 & Top level domain length & \xmark & \cmark \\
        4 & Domain Unique Characters length & \xmark & \cmark \\
        5 & SLD Unique Characters length & \xmark & \cmark \\
        6 & TLD Unique Characters length & \xmark & \cmark \\
        7 & Has malicious TLD & \xmark & \cmark \\
        8 & Has Valid TLD & \cmark & \xmark \\
        9 & TLD Hash & \xmark & \cmark \\
        10 & Contains Digits & \cmark & \xmark \\
        11 & Starts with Digit & \xmark & \cmark \\
        12 & Underscore Ratio* & \cmark & \xmark \\
        13 & Symbol ratio & \xmark & \cmark \\
        14 & Hex ratio & \xmark & \cmark \\
        15 & Digit Ratio* & \cmark & \cmark \\
        16 & Vowel Ratio* & \cmark & \cmark \\
        17 & Consonant Ratio &\xmark & \cmark \\
        18 & Ratio of Repeated Characters* & \cmark & \cmark \\
        19 & Ratio of Consecutive Consonants* & \cmark & \cmark \\
        20 & Ratio of Consecutive Digits* & \cmark & \cmark \\
        21 & Number of tokens in SLD & \xmark & \cmark \\
        22 & Number of digits in SLD & \xmark & \cmark \\
        23 & Entropy* & \cmark & \cmark \\
        24 & Gini Index & \xmark & \cmark \\
        25 & Classification error of characters & \xmark & \cmark \\
        26 & N-Gram Distribution* & \cmark &\xmark \\
        27 & 2-Gram Median & \xmark & \cmark \\
        28 & 3-Gram Median & \xmark & \cmark \\
        29 & 2-Gram Circle Median & \xmark & \cmark \\
        30 & 3-Gram Circle Median & \xmark & \cmark \\
        31 & Number of Subdomains** & \cmark & \xmark \\
        32 & Subdomain Length Mean** & \cmark & \xmark \\
        33 & Has www Prefix & \cmark & \xmark \\
        34 & Contains Single-Character Subdomain** & \cmark & \xmark \\
        35 & Is Exclusive Prefix Repetition & \cmark & \xmark \\
        36 & Contains TLD as Subdomain** & \cmark & \xmark \\
        37 & Ratio of Digit-Exclusive Subdomains** & \cmark & \xmark \\
        38 & Ratio of Hexadecimal-Exclusive Subdomains** & \cmark & \xmark \\
        39 & Contains IP Address** & \cmark & \xmark \\
        40 & Alphabet Cardinality* & \cmark & \xmark \\
        \bottomrule
    \end{tabular}
    \caption{Features used by FANCI and B-RF. (*) For these features, FANCI uses dot free public-suffix-free domain. (**) For these features, FANCI uses public-suffix-free domain.}
    \label{tab:features}
\end{table}

\subsection{LSTM.MI}
Woodbridge et al.~\cite{Woodbridge2016} were the first to propose deep learning for DGA domain name detection. Their DGA classifier is a neural network consisting of an embedding layer, an LSTM layer, and a single node output layer with sigmoid activation. In this paper, we use the LSTM.MI model that was proposed recently by Tran et al.~\cite{tran2018lstm}. Its architecture is very similar to that of Woodbridge et al.~\cite{Woodbridge2016}; the main distinction is that the LSTM.MI model is trained with a cost-sensitive learning algorithm that takes class imbalances into account. This allows the LSTM.MI approach to achieve slightly better results than the original LSTM approach (see \cite{tran2018lstm,rsivaguru2018}). The code for training the LSTM.MI model is publicly available.\footnote{\url{https://github.com/bkcs-hust/lstm-mi}. Accessed: 2019-02-08.}

\subsection{FANCI}
The FANCI classifier recently proposed by Sch\"uppen et al.~\cite{schuppen2018} is a random forest (RF) classifier designed to classify NXDomains as benign (bNXD) or malicious (mAGD). NXDomains, or Non-Existent Domains, are domains that can not be resolved. DGAs generate hundreds or even thousands of domains every day, only very few of which are actually registered by the botmaster. That means that almost all queries for DGA generated domains by infected machines will result in an NXDomain response by the local DNS server, so it is reasonable to attempt to detect DGA activity by analyzing NXDomains.

To this end, the FANCI classifier leverages 21 manually defined features, extracted from the domain name string. The 21 features can be divided into structural, linguistic, and statistical categories (see \cref{tab:features}). The FANCI RF model is comprised of 9 decision trees, of which 7 use the Gini coefficient as the measure of impurity and the other 2 use entropy. Each tree takes between 2 to 18 features. The source code of the FANCI classifier is available on GitHub.\footnote{\url{https://github.com/fanci-dga-detection/fanci}. Accessed: 2019-02-08.}

The domain names used in our experiments contain only SLDs and TLDs (see section \ref{sec:data}).
As such, it is expected that a number of features used in the FANCI model would not make a distinction between malicious and benign examples. \Cref{tab:unused FANCI features} lists the FANCI features that are not expected to have any effect.

\begin{table}
    \centering
    \begin{tabular}{cl}
        \toprule
        \# & Feature  \\
        \midrule
        31 & Number of Subdomains  \\
        33 & Has www Prefix  \\
        34 & Contains Single-Character Subdomain \\
        35 & Is Exclusive Prefix Repetition \\
        36 & Contains TLD as Subdomain \\
        \bottomrule
    \end{tabular}
    \caption{FANCI features not expected to have any effect in our experiments}
    \label{tab:unused FANCI features}
\end{table}

\subsection{B-RF}
B-RF~\cite{rsivaguru2018} is a random based DGA detection classifier that is trained on 26 manually engineered features as indicated in \cref{tab:features}. There is some overlap between the features used by FANCI and those used by B-RF. For instance, both make use of the domain name length, digit and vowel ratio, ratio of repeated characters, etc. Some features are used by FANCI but not by B-RF, such as whether the domains have valid TLDs or whether they contain digits. Other features like 2-gram median and 3-gram median are only used by B-RF.

B-RF consists of 100 trees and each tree is trained using a subset with a maximum of 20 features. Entropy is used as the criterion to decide the split attribute while growing the trees in the random forest.


\subsection{Results}\label{sec:results}

\begin{table*}
    \centering
    \begin{tabular}{llrrrr}
        \toprule
        \multirow{2}{*}{Classifier} & \multirow{2}{*}{Data set} & \multicolumn{2}{c}{FPR=0.001} & \multicolumn{2}{c}{FPR=0.01}\\
        \cmidrule(lr){3-4} \cmidrule(lr){5-6}
        & & TPR & AUC & TPR & AUC\\
        \midrule
         \multirow{8}{*}{LSTM.MI}
         & AlexaBamb & 96.79\% & 94.91\% & 99.27\% & 98.89\%\\
         & AlexaBamb + CharBot & 95.50\% & 95.35\% & 98.89\% & 98.67\%\\
         & AlexaBamb + DeepDGA & 96.65\% & 96.44\% & 99.20\% & 99.00\%\\
         & AlexaBamb + DeceptionDGA & 95.17\% & 95.05\% & 98.54\% & 98.46\%\\
         \cmidrule{2-6}
         & QnameBamb & 81.98\% & 83.37\% & 98.98\% & 96.68\%\\
         & QnameBamb + CharBot & 82.91\% & 84.98\% & 98.51\% & 96.48\%\\
         & QnameBamb + DeepDGA & 83.22\% & 83.98\% & 98.85\% & 96.50\%\\
         & QnameBamb + DeceptionDGA & 84.66\% & 85.57\% & 98.61\% & 96.82\%\\
         \midrule
         \multirow{8}{*}{FANCI}
         & AlexaBamb & --- & --- & 74.46\% & 80.46\%\\
         & AlexaBamb + CharBot & --- & --- & 72.49\% & 78.71\%\\
         & AlexaBamb + DeepDGA & --- & --- & 73.98\% & 80.02\%\\
         & AlexaBamb + DeceptionDGA & --- & --- & 73.84\% & 80.18\%\\
         \cmidrule{2-6}
         & QnameBamb & --- & --- & 74.13\% & 79.06\%\\
         & QnameBamb + CharBot & --- & --- & 72.13\% & 77.80\%\\
         & QnameBamb + DeepDGA & --- & --- & 72.89\% & 78.19\%\\
         & QnameBamb + DeceptionDGA & --- & --- & 73.65\% & 78.86\%\\
         \midrule
         \multirow{8}{*}{B-RF}
         & AlexaBamb & 85.72\% & 82.93\% & 94.72\% & 94.67\%\\
         & AlexaBamb + CharBot & 84.62\% & 79.30\% & 93.89\% & 93.75\%\\
         & AlexaBamb + DeepDGA & 85.81\% & 80.94\% & 94.39\% & 94.35\%\\
         & AlexaBamb + DeceptionDGA & 84.51\% & 78.62\% & 93.78\% & 93.73\%\\
         \cmidrule{2-6}
         & QnameBamb & 82.75\% & 73.85\% & 96.88\% & 94.52\%\\
         & QnameBamb + CharBot & 83.03\% & 74.98\% & 96.37\% & 94.29\%\\
         & QnameBamb + DeepDGA & 82.64\% & 73.87\% & 96.39\% & 94.26\%\\
         & QnameBamb + DeceptionDGA & 82.98\% & 76.26\% & 96.26\% & 94.56\%\\
         \bottomrule
    \end{tabular}
    \caption{Performance metrics of LSTM.MI, FANCI and B-RF on the different data sets.}
    \label{tab:results}
\end{table*}

\begin{table*}
    \centering
    \begin{tabular}{llrrrrrr}
        \toprule
         \multirow{2}{*}{Classifier} & \multirow{2}{*}{Data set} & \multicolumn{3}{c}{FPR=0.001} & \multicolumn{3}{c}{FPR=0.01}\\
         \cmidrule(lr){3-5} \cmidrule(lr){6-8}
         & & CharBot & DeepDGA & DeceptionDGA & CharBot & DeepDGA & DeceptionDGA\\
         \midrule
         \multirow{8}{*}{LSTM.MI}
         & AlexaBamb & 5.58\% & 33.98\% & 4.02\% & 15.50\% & 39.53\% & 12.74\%\\
         & AlexaBamb + CharBot & 55.19\% & 92.54\% & 19.69\% & 81.08\% & 98.44\% & 47.67\%\\
         & AlexaBamb + DeepDGA & 12.39\% & 98.35\% & 7.34\% & 12.39\% & 98.35\% & 7.34\%\\
         & AlexaBamb + DeceptionDGA & 23.59\% & 88.71\% & 40.29\% & 52.18\% & 96.66\% & 71.52\%\\
         \cmidrule{2-8}
         & QnameBamb & 15.25\% & 6.18\% & 16.61\% & 31.90\% & 19.51\% & 37.73\%\\
         & QnameBamb + CharBot & 52.67\% & 42.48\% & 34.45\% & 81.96\% & 85.90\% & 66.27\%\\
         & QnameBamb + DeepDGA & 27.84\% & 94.51\% & 24.25\% & 43.28\% & 99.61\% & 47.33\%\\
         & QnameBamb + DeceptionDGA & 30.45\% & 15.97\% & 37.74\% & 53.31\% & 37.97\% & 24.25\%\\
         \midrule
         \multirow{8}{*}{FANCI}
         & AlexaBamb & --- & --- & --- & 3.05\% & 6.33\% & 1.66\%\\
         & AlexaBamb + CharBot & --- & --- & --- & 22.26\% & 12.12\% & 2.64\%\\
         & AlexaBamb + DeepDGA & --- & --- & --- & 6.45\% & 83.17\% & 2.08\%\\
         & AlexaBamb + DeceptionDGA & --- & --- & --- & 4.27\% & 6.57\% & 2.33\%\\
         \cmidrule{2-8}
         & QnameBamb & --- & --- & --- & 21.43\% & 5.37\% & 46.85\%\\
         & QnameBamb + CharBot & --- & --- & --- & 48.44\% & 14.20\% & 49.62\%\\
         & QnameBamb + DeepDGA & --- & --- & --- & 45.13\% & 77.88\% & 50.11\%\\
         & QnameBamb + DeceptionDGA & --- & --- & --- & 44.75\% & 13.77\% & 50.45\%\\
         \midrule
         \multirow{8}{*}{B-RF}
         & AlexaBamb & 1.69\% & 6.61\% & 1.38\% & 27.59\% & 23.97\% & 22.37\%\\
         & AlexaBamb + CharBot & 1.84\% & 9.19\% & 1.20\% & 64.33\% & 34.06\% & 41.22\%\\
         & AlexaBamb + DeepDGA & 4.54\% & 46.55\% & 2.86\% & 31.80\% & 84.12\% & 26.86\%\\
         & AlexaBamb + DeceptionDGA & 2.00\% & 8.23\% & 1.34\% & 33.14\% & 24.51\% & 32.53\%\\
         \cmidrule{2-8}
         & QnameBamb & 18.80\% & 2.99\% & 41.67\% & 61.05\% & 22.57\% & 62.31\%\\
         & QnameBamb + CharBot & 43.47\% & 12.54\% & 49.18\% & 85.82\% & 60.70\% & 79.33\%\\
         & QnameBamb + DeepDGA & 44.04\% & 33.08\% & 49.80\% & 65.97\% & 98.84\% & 67.79\%\\
         & QnameBamb + DeceptionDGA & 39.97\% & 10.53\% & 47.86\% & 62.29\% & 23.30\% & 67.74\%\\
         \bottomrule
    \end{tabular}
    \caption{Detection rates of the different DGAs.}
    \label{tab:rates}
\end{table*}

\begin{figure}
    \centering\includegraphics[width=.45\textwidth]{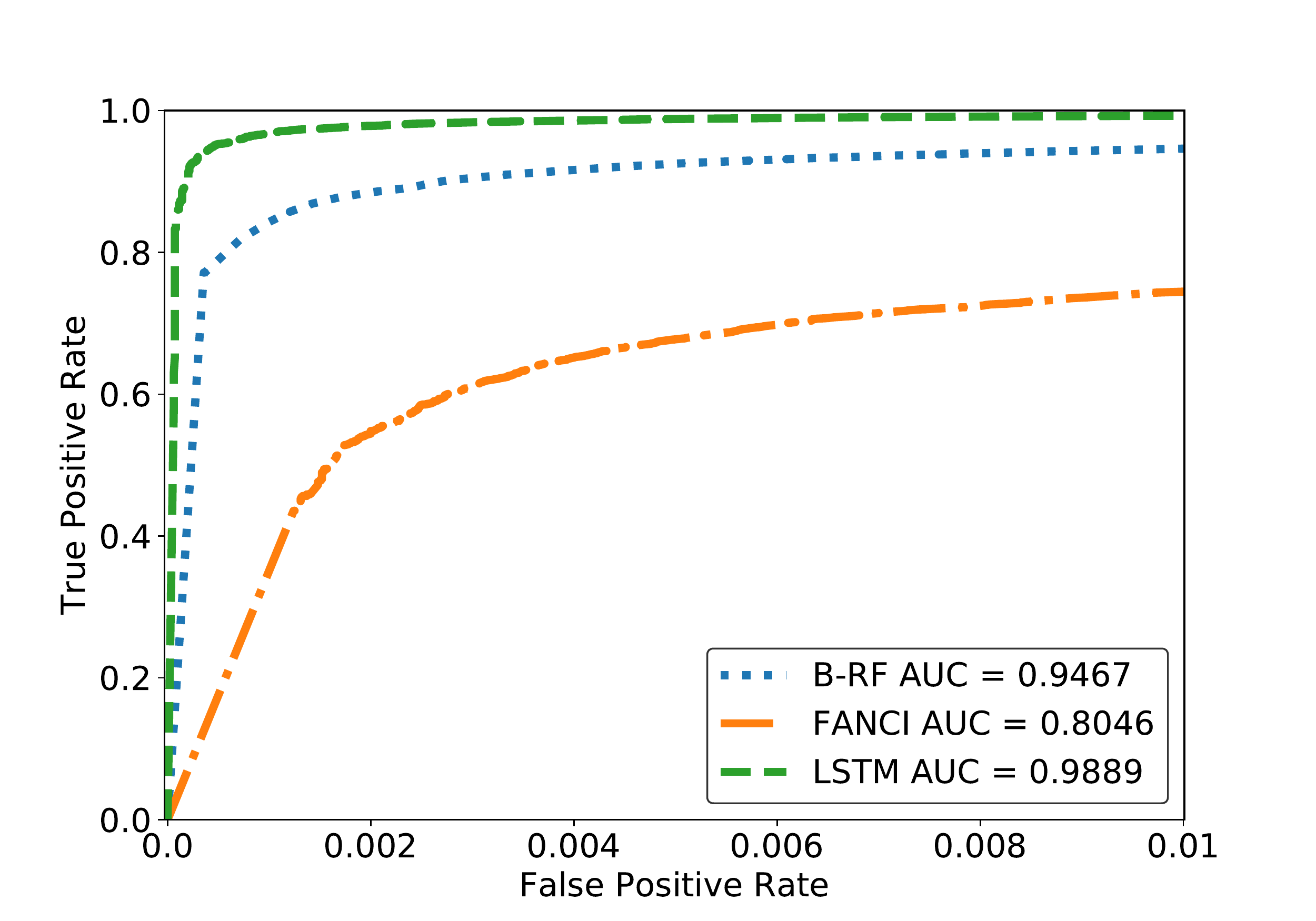}
    \caption{ROC curves for the classifiers trained on the AlexaBamb data set.}
    \label{fig:roc_curve}
\end{figure}

The predictive performance metrics are summarized in \cref{tab:results} for false positive rates\footnote{One can argue that even a FPR of 0.1\% is still too high to be useful in practice. While this can certainly be true depending on the application, note that lower FPRs can only make our results better as the models will necessarily have lower TPR and lower detection rates for CharBot.} of 0.1\% and 1\%. \Cref{fig:roc_curve} shows ROC curves for the different models on the AlexaBamb data. We plot the ROC curve only for FPRs between 0 and 0.01, as higher FPRs are meaningless in practice. We conclude from these results that the deep learning approach does better than the RF approaches, which is in line with what has been reported before in the literature \cite{Woodbridge2016, Yu2018a, Yu:2017a, lison2017automatic, tran2018lstm}. Among the RF models, B-RF outperforms FANCI significantly. We found that this improvement was not due to the number of trees, as decreasing the number of trees used by B-RF from 100 to 9 (as in FANCI) still yielded superior performance for B-RF. We therefore believe this difference in performance is caused by the different feature sets.

We were unable to establish a classification threshold that achieves 0.1\% FPR for FANCI. Therefore, in reporting FANCI results, we only consider FPR = 1\%. We believe this is due to the fact that \cite{schuppen2018} used proprietary data to filter classification outcomes which increased accuracy. Our use of a different post-filtering data set may be the reason for the difference (note that the authors of the FANCI paper use accuracy whereas we use AUC). For AlexaBamb and its augmented datasets, we obtained between 90.7\% and 91.4\% accuracy. For QnameBamb and its augmented datasets, we obtained between 92.5\% and 93.16\% accuracy. This is not too far from the 93.7\% that was reported in \cite{spooren2019detection} where the authors also attempted to replicate the FANCI results.

All models fail to adequately detect CharBot and DeceptionDGA domains even when explicitly trained on them. The LSTM.MI model succeeds in detecting DeepDGA close to 99\% of the time with adversarial training, but the other models generally fail at detecting DeepDGA as well. Training on Qname instead of Alexa makes a significant difference, both in predictive performance as well as detection rate: the models have lower predictive accuracy when trained on Qname, but they are better able to detect CharBot domains. At the 0.1\% FPR, however, these detection rates are nowhere near high enough to be useful in practice. FANCI is unable to properly detect CharBot at 1\% FPR, whereas LSTM.MI and B-RF sometimes manage to obtain over 80\% detection rate here. This is not a very useful result, however, since 1\% FPR is considered too high to be practical. Therefore, at a low FPR, the domains generated by CharBot can be said to be \textit{transferable} across different models and data sets in the sense that CharBot can fool models that have vastly different architectures and are not trained on Alexa. Combined with its simplicity, speed and small size, this makes CharBot an ideal DGA for use in malware in the wild.

\begin{figure*}
    \centering\includegraphics[width=\textwidth]{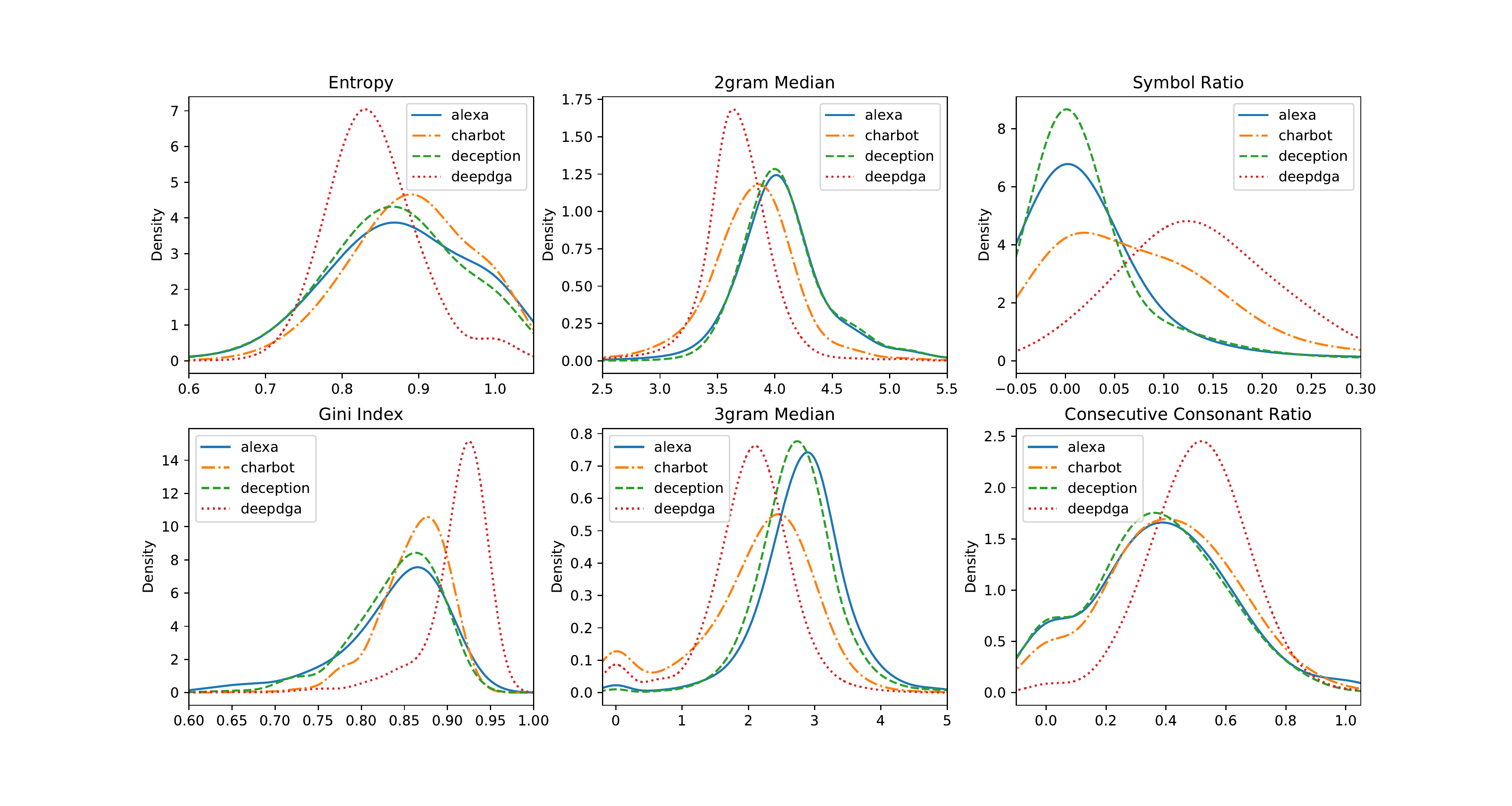}
    \caption{Kernel density estimation of the feature distributions of Alexa and adversarial domain names.}
    \label{fig:feature_analysis}
\end{figure*}

The success of CharBot may be explained as follows. The algorithm works by taking the Alexa list of benign domains --- which most DGA classifiers would overwhelmingly classify as such --- and introduces a small number of typographical errors. The statistical properties of the domains generated by CharBot are therefore likely almost identical to those of Alexa, causing a low detection rate. The transferability may be explained by noting that even though Alexa and Qname are different data sets, they still capture the same underlying distribution: namely, that of benign domains. This closeness in distribution is most likely shared among all sufficiently large corpora of benign domains, allowing CharBot to fool any DGA classifier that only takes the domain name string into account. We test this hypothesis by performing kernel density estimation on the feature distributions of the Alexa domains and the adversarial domains. The results are plotted in \cref{fig:feature_analysis}. The Entropy and Gini index features are standard impurity measures for decision trees. The other features are:
\begin{itemize}
    \item 2gram Median. This feature takes the median frequency from the list of 2gram frequencies for the given SLD. Bigram frequencies are collected from the Python package called wordfreq \footnote{\url{https://pypi.org/project/wordfreq/1.1/}. Accessed: 2019-02-14.}. 
    
    \item 3gram Median. 3gram median is similar to 2gram median except that it returns the median frequency from the list of trigram frequencies for the given SLD.
    
    \item Symbol ratio. This feature defines the ratio of non-alphabetical characters in the SLD, which includes digits and special characters.
    
    \item Consecutive Consonant Ratio. This feature defines the ratio of consecutive consonants in the SLD.
\end{itemize}
From the plots, we observe that the feature distributions of CharBot domains are much closer to those of Alexa than the distributions of DeepDGA are. However, DeceptionDGA is more similar to Alexa than CharBot is, although the difference is very small in some cases. Nevertheless, CharBot gets quite close to Alexa, which explains why it is so successful in fooling DGA classifiers. It also shows that defending against CharBot may be very difficult, potentially requiring a very high FPR. \Cref{fig:feature_analysis} also provides insights into what parts of CharBot may be improved to yield an even more effective DGA:
\begin{itemize}
    \item The entropy curve of CharBot can be made more similar to that of Alexa domains by using a different replacement character distribution. Currently, we are using the uniform distribution, which has the highest possible entropy. Switching to a lower entropy distribution may improve the performance of CharBot, although this would need to be carefully balanced against the probability of a generated domain already being registered.

    \item The random replacement of two characters caused the 2-gram distributions of CharBot to differ from those of the Alexa domains. We can overcome this weakness by replacing neighboring characters with 2-grams that occur frequently in Alexa. A similar line of reasoning applies to 3-grams.

    \item The symbol ratio distributions can be made more similar by drawing $c_1$ and $c_2$ from the same letters or digit sets of the original domain. For example, replace a digit with another random digit and not with a letter.
\end{itemize}

Investigating the lengths of the domain names that were generated vs. those that are present in the Alexa and Qname data sets (see \cref{tab:stats}), we find that CharBot names are close in length to Alexa names (which is to be expected), but Qname, Bambenek and DeepDGA domains are significantly longer on average, whereas DeceptionDGA are significantly shorter. This difference in lengths may contribute to the detection rates: when training on Alexa, CharBot domains are similar in length whereas DeepDGA domains are longer like the Bambenek domains. By contrast, when training on Qname, domains are longer on average, which aids detection of CharBot (although the difference is not very large).

\section{Countermeasures}\label{sec:counter}
We consider a few options for defending against attacks such as CharBot:

\subsection{Comparing incoming domains to Alexa}
The simplest defense against CharBot would be to take the domain in question and compare it to the full Alexa list. If the domain is equal to one found in the Alexa list save for one or two replaced characters, the domain is flagged as malicious. However, the Alexa data set contains one million samples, so this approach of computing the Hamming distance of input domains on the fly may not be practical. We can make this computation even harder by modifying CharBot to perform deletions and insertions, forcing the use of the edit distance~\cite{levenshtein1966binary} rather than the Hamming one. Practical implementations can reduce lookup time by pre-computing noisy versions of the Alexa list into a compact data structure such as a Bloom filter~\cite{bloom1970space}. However, this approach is marred by a combinatorial explosion of possible corrupted domain names based on the Alexa data set: if we let $n$ be the size of the CharBot data set, $\ell$ be the average length of a domain name, $k$ be the number of edits CharBot introduces and $m$ be the size of the replacement alphabet, then the number of possible domains CharBot can generate is given approximately by
$$
    n{\ell \choose k}(m-1)^k.
$$
For $n = 10,000$, $\ell = 16$, $m = 40$ and $k = 2$ this yields 1,825,200,000 possible domains. Besides, this defense can also easily be defeated by simply using a different legitimate data set instead of Alexa for generating domain names.

\subsection{Increasing the capacity of the models}
Using more complicated classification models may allow them to find a meaningful separation between Alexa and CharBot domains. However, this would require careful feature engineering for featureful models and increase the computational burden of both model training and inference. Given that practical DGA classifiers need to be regularly retrained to keep up with new malware and they need to process many domains in real-time, this may not be feasible. Nevertheless, this may be an option worth exploring in future work.

\subsection{White-box adversarial training}
Our adversarial training procedure in this paper has consisted of generating a list of adversarial domains once and then augmenting the training data with them. However, adversarial training is usually done iteratively: at every iteration of training, the current batch of training samples is augmented with adversarially generated set specifically for the model at that particular stage~\cite{goodfellow2014explaining,madry2017towards}. This requires a \textit{white-box} attack which is able to take the model parameters into account. Adversarial attacks have mostly been considered in the image domain, although there is some work on text classification~\cite{gong2018adversarial,ebrahimi2018hotflip}. Making use of this recent body of work on white-box adversarial training for text classification may allow us to improve the detection rate of CharBot.

\subsection{Using side information}
Perhaps the most realistic defense against attacks like CharBot would be to use additional information besides the domain name string alone. For instance, the IP addresses the domain maps to, how often the domain was queried and when, etc. There have been several works investigating the use of such information in DGA classification~\cite{yadav2012detecting,schiavoni2014phoenix,kwon2016psybog,lison2017neural,pereira2018a}. A fruitful avenue for future work could be to test whether these classifiers are more resilient to CharBot.

\begin{table}
    \centering
    \begin{tabular}{lrr}
        \toprule
        Data Set & Mean & Standard Deviation\\
        \midrule
        Alexa & 14.30 & 4.70\\
        Bambenek & 21.80 & 6.42\\
        Qname & 23.99 & 7.97\\
        CharBot & 14.27 & 4.02\\
        DeepDGA & 28.52 & 9.06\\
        DeceptionDGA & 10.20 & 3.90\\
        \bottomrule
    \end{tabular}
    \caption{Statistics of the domain name lengths for each data set.}
    \label{tab:stats}
\end{table}

\section{Conclusion}\label{sec:conclusion}
We have proposed CharBot, a simple and efficient DGA. We have shown CharBot to be effective at both generating large amounts of unregistered domain names as well as fooling three DGA classifiers: FANCI, LSTM.MI and B-RF. We also compared CharBot to DeepDGA and DeceptionDGA, two state-of-the-art domain generation algorithms. The domain names generated by CharBot were more likely to be unregistered than those generated by DeepDGA or DeceptionDGA. Moreover, adversarial retraining using CharBot, DeepDGA or DeceptionDGA did not result in adequate detection of CharBot domains names.

Our DGA is the very first example of a black-box adversarial machine learning attack against DGA classifiers that is not based on Generative Adversarial Networks. We show that simply introducing small perturbations to a set of legitimate domains is good enough and such advanced techniques are unnecessary. We believe this highlights a dangerous weakness of modern DGA classifiers, namely their vulnerability to extremely simple attacks that make no use of sophisticated machine learning techniques. CharBot is an algorithm that could be realistically used in malware in the wild to circumvent state of the art DGA classifiers, making it a real threat. We speculate that this vulnerability is actually \textit{inherent} to any classifier that relies only on the domain name string to perform DGA classification. The CharBot DGA is similar to dictionary DGAs: both have a list of strings embedded as part of the DGA code. In the case of dictionary DGAs this list is a dictionary of words that are combined in various ways to generate a domain name, while in the case of CharBot the list contains benign domain names that are altered slightly to generate a new domain name for malicious purposes. In both cases, the generated domain names exhibit properties that are very close to natural language, which makes them extremely difficult to distinguish from benign domain names.

Machine learning models that attempt to do DGA classification based only on the domain name itself, such as the ones considered in this paper, might not be sufficient to detect a DGA like CharBot. The result highlights the need for ML models that exploit additional context features such as the IP-addresses that the domains are mapped to, or temporal access patterns (e.g.~how often the domain was requested, and when)~\cite{yadav2012detecting,schiavoni2014phoenix,kwon2016psybog,lison2017neural}, as was done successfully for dictionary DGAs~\cite{pereira2018a}.

For future work, we focus on defending DGA classifiers against simple attacks such as CharBot. The avenues we are investigating to achieve this include performing white-box adversarial training as well as augmenting the model inputs with side information that is more difficult to manipulate.

\section*{Reproducibility}
To foster reproducibility of our results, we are open to sharing all of our code as well as data sets of CharBot samples upon request.

\section*{Acknowledgement}
We gratefully acknowledge the support of NVIDIA Corporation with the donation of the Titan Xp GPU used for this research. We thank Bobby Filar for making the code of the original DeepDGA algorithm available to us~\cite{anderson2016deepdga} and Jan Spooren for providing us with domain names generated by DeceptionDGA~\cite{spooren2019detection}. Jonathan Peck is sponsored by a Ph.D. fellowship from the Research Foundation Flanders (FWO).

\bibliographystyle{IEEEtran}
\bibliography{references}

\end{document}